\documentclass{article}
\PassOptionsToPackage{numbers, compress}{natbib}
\usepackage[main, final]{neurips_2026}
\usepackage[utf8]{inputenc} % allow utf-8 input
\usepackage[T1]{fontenc}    % use 8-bit T1 fonts
\usepackage{hyperref}       % hyperlinks
\usepackage{url}            % simple URL typesetting
\usepackage{booktabs}       % professional-quality tables
\usepackage{amsfonts}       % blackboard math symbols
\usepackage{nicefrac}       % compact symbols for 1/2, etc.
\usepackage{microtype}      % microtypography
\usepackage{xcolor}         % colors
\usepackage{graphicx}
\usepackage{amsmath}

\title{StatLUT: Statistical Feature-Driven Multimodal 3DLUT Generation for Photorealistic Style Transfer}

\author{
  Yifan Wang, \quad Zhixiang Hao, \quad Yu Wang\thanks{Corresponding author.}, \quad Congchao Zhu \\
  Honor Device Co., Ltd. \\
  \texttt{wyf141060@163.com, \{haozhixiang, wangyu24, zhucongchao\}@honor.com} \\
}

\begin{document}

\maketitle

\begin{abstract}
    Photorealistic Style Transfer (PST) aims to transfer the color and tonal style of a reference to a content image while strictly preserving its structural integrity. However, existing deep learning-based methods inherently suffer from semantic entanglement caused by pre-trained image encoders, leading to unnatural spatial distortions. Moreover, current pixel-level mapping paradigms often ignore color gamut topology, resulting in color banding, while also lacking the multimodal capability for precise, tone-specific text-driven control. To address these bottlenecks, we propose StatLUT, an innovative multimodal framework for 3D LUT generation. First, we bypass traditional encoders and introduce a Lab-Extractor to derive spatially-agnostic statistical features, fundamentally decoupling color distributions from structural semantics to ensure artifact-free rendering. Second, we formulate LUT generation as a Transformer-based Seq2Seq translation task, utilizing a Multi-dimensional Residual Mapper (MR-Mapper) to predict topologically smooth 3D LUTs. Finally, to break the unimodal barrier, we propose the H-Diffuser, a lightweight Diffusion Transformer that directly synthesizes statistical features from tone-descriptive text prompts, enabling controllable and robust color grading. Extensive experiments on standard benchmarks demonstrate that StatLUT significantly outperforms state-of-the-art methods in both visual quality and quantitative metrics, pioneering a highly robust and versatile paradigm for multimodal photorealistic style transfer.
\end{abstract}

\section{Introduction}
\label{sec:intro}

Photorealistic Style Transfer (PST) aims to faithfully transfer the color and tonal style of a reference image to a content image while strictly preserving the original spatial structure. Unlike artistic style transfer\cite{anUltrafastPhotorealisticStyle2020, gatysImageStyleTransfer2016, huangArbitraryStyleTransfer2017, johnsonPerceptualLossesRealTime2016, parkArbitraryStyleTransfer2019}, PST demands visual realism and structural fidelity, precluding unnatural artifacts or texture distortions during non-linear color mapping.

Although deep learning has advanced PST, adhering to this ``artifact-free'' baseline remains challenging. Current mainstream methods widely adopt pre-trained encoder-decoder architectures\cite{liClosedformSolutionPhotorealistic2018, yooPhotorealisticStyleTransfer2019, chiuPhotoWCT2CompactAutoencoder2022}. However, these models naturally extract high-level semantic features, creating a fundamental mechanistic mismatch with the PST task, which inherently relies on low-level color distributions. This semantic entanglement allows local structural semantics to interfere with color mapping, leading to spatial distortions. Furthermore, their massive memory consumption constitutes a prohibitive bottleneck for high-resolution (e.g., 4K/8K) deployment\cite{linAdaCMAdaptiveColorMLP2023, liDLUTPhotorealisticStyle2025}.

To overcome these barriers, recent research has shifted towards a pixel-level mapping paradigm\cite{linAdaCMAdaptiveColorMLP2023, liDLUTPhotorealisticStyle2025, keNeuralPresetColor2023}, with 3D lookup tables (LUTs) emerging as the standard due to their non-linear capacity and low computational cost\cite{liDLUTPhotorealisticStyle2025}. Nevertheless, existing LUT generation mechanisms exhibit significant limitations: 1) traditional blending paradigms\cite{chenNLUTNeuralbased3D2023, gongSALUTSpatialAdaptive2025} restrict representations to a linear subspace, hindering out-of-distribution generalization; 2) recent generative paradigms\cite{shinVideoColorGrading2025} fail to completely decouple semantics from color, risking texture artifacts; 3) point-wise mapping approaches\cite{linAdaCMAdaptiveColorMLP2023} ignore the topology between adjacent color bins, causing color banding and inter-frame flickering in videos. Moreover, they lack multimodal capabilities for precise, tone-specific text-driven color control.

To address these limitations, we propose StatLUT, a novel multimodal photorealistic style transfer and color generation framework. In summary, our main contributions are as follows:

\begin{itemize}
    \item Bypassing traditional image encoders, we utilize multi-dimensional low-level statistical distributions to construct explicit color priors. This blocks semantic interference during color mapping, preserving structural fidelity and mitigating spatial distortions.

    \item By adopting a Transformer-based sequence-to-sequence paradigm, our method globally accounts for the mesh topology of the 3D LUT. This generates smooth color manifolds, which effectively resolves the issues of color banding and temporal flickering.

    \item We introduce a Diffusion Transformer (DiT) architecture\cite{peeblesScalableDiffusionModels2023} to break the unimodal barrier, enabling color control via tone-descriptive text prompts without relying on reference images, thereby expanding the application scope of PST.
\end{itemize}

\section{Related Work}
\label{sec:related}

\subsection{Feature-Encoding-based Photorealistic Style Transfer}
Unlike artistic style transfer, PST strictly preserves structural integrity. Early optimization-based methods\cite{liClosedformSolutionPhotorealistic2018, pitieNdimensionalProbabilityDensity2005} achieved this via photorealistic regularization but suffered from slow iterative processes. Subsequent feed-forward approaches and AdaIN-based pipelines\cite{huangArbitraryStyleTransfer2017} accelerated this process by utilizing pre-trained vision models for feature fusion.

However, this ``Encode-Fuse-Decode'' paradigm has two fundamental limitations. First, pre-trained encoders inherently extract high-level semantics\cite{simonyanVeryDeepConvolutional2015, tanEfficientnetRethinkingModel2019}, causing a mechanistic mismatch with PST. Consequently, relying solely on loss functions fails to disentangle semantics from color, inevitably leading to texture artifacts. Second, processing full-resolution feature maps via heavy decoders incurs prohibitive computational costs, making real-time high-resolution rendering unattainable.

\subsection{Pixel-Level Mapping and 3D LUT Generation}
To bypass decoder bottlenecks, researchers explored pixel-level mapping. Early statistical and matrix-based methods\cite{reinhardColorTransferImages2001, kangPhotorealisticImageStyle2024} were extremely fast but lacked complex non-linear expressiveness. Bilateral grids\cite{xiaJointBilateralLearning2020} introduced local mapping but disrupted inter-frame consistency in videos.

Consequently, 3D LUTs emerged as the optimal solution. However, basis-LUT blending paradigms\cite{chenNLUTNeuralbased3D2023, gongSALUTSpatialAdaptive2025} restrict representations to a linear subspace, resulting in poor out-of-distribution generalization. Recent alternatives also face significant challenges: point-wise mappings ignore color lattice topology, causing banding, and reliance on pre-trained features inevitably introduces semantic interference\cite{linAdaCMAdaptiveColorMLP2023, karTemporalConsistentSemantic2025}. Diffusion-based methods require expensive instance-specific retraining\cite{liDLUTPhotorealisticStyle2025}, while other generative approaches\cite{shinVideoColorGrading2025} still mix structural semantics into condition vectors, failing to eliminate the risk of semantic entanglement.

\subsection{Multimodal and Text-Driven Image Manipulation}
While text-guided image manipulation has advanced rapidly\cite{fuGuidingInstructionbasedImage2023, kawarImagicTextBasedReal2023, brooksInstructPix2PixLearningFollow2023, radfordLearningTransferableVisual2021, mengSDEditGuidedImage2021}, most works focus on generation\cite{mengSDEditGuidedImage2021} or colorization\cite{changLCADLanguagebasedColorization2023}. Colorization fundamentally differs from PST: it acts as a ``painter'' filling colors into grayscale structures, whereas PST acts as a ``colorist'' establishing precise color mappings to alter stylistic atmospheres.

In PST, multimodal exploration remains scarce. Existing text-to-image pipelines\cite{kawarImagicTextBasedReal2023, brooksInstructPix2PixLearningFollow2023} often overlook the fact that PST intrinsically relies on color priors, not specific semantics (e.g., a ``sunset beach'' and a ``twilight forest'' share similar colors despite different semantics). A direct, efficient, and temporally consistent text-to-LUT mechanism remains an unresolved challenge.

\section{Methodology}
\label{sec:method}

\begin{figure}[t]
    \centering
    \includegraphics[width=1\linewidth]{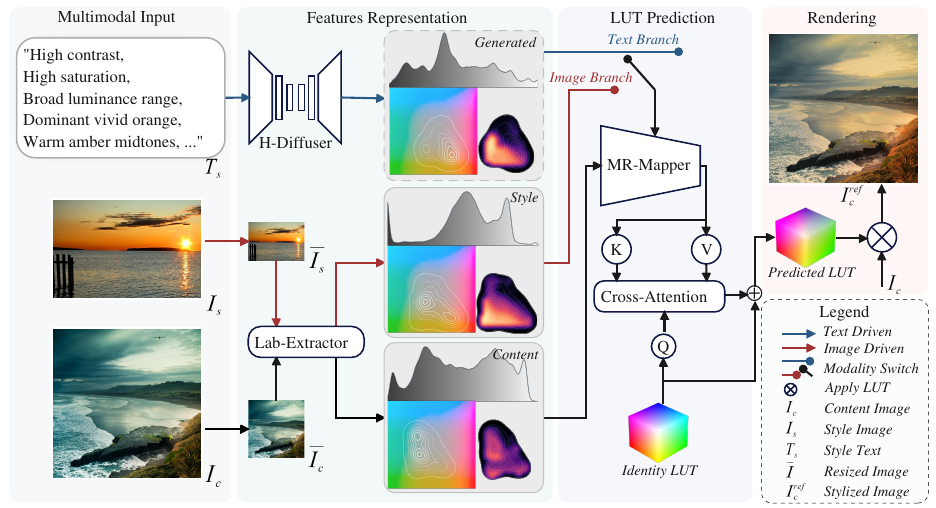}
    \caption{Overview of the proposed StatLUT framework. (1) Feature Extraction: The Lab-Extractor extracts spatially-agnostic statistical features from resized input images. (2) LUT Prediction: The MR-Mapper fuses these features to guide a Cross-Attention module, which transforms an identity LUT into the predicted 3D LUT. (3) Text-Driven Mapping: For text conditions, the H-Diffuser directly generates target statistics. Finally, the predicted LUT is applied to stylize the high-resolution content image.}
    \label{fig:framework}
\end{figure}

Given a high-resolution content image $I_c$ and a style condition (reference image $I_s$ or text prompt $T_s$), StatLUT predicts a 3D LUT for color stylization, transforming the source colors of $I_c$ to align with the target style distribution (Figure \ref{fig:framework}). To eliminate semantic entanglement, we extract spatially-agnostic statistical features (Sec. \ref{sec:decoupling}). These features guide a Seq2Seq cross-attention module to predict a topologically smooth 3D LUT (Sec. \ref{sec:prediction}). Finally, a DiT-based module enables text-driven color generation by directly synthesizing target statistics (Sec. \ref{sec:text-driven}).

\subsection{Spatially-Agnostic Feature Extraction via Lab-Extractor}
\label{sec:decoupling}

Since the 3D LUT formulation inherently guarantees the structural preservation of the content, the primary challenge in PST is extracting pure color and illumination distributions from the style image while strictly discarding spatial or semantic content. Existing approaches relying on deep encoders \cite{liuInstantPhotorealisticStyle2023} or AdaIN \cite{huangArbitraryStyleTransfer2017} inevitably suffer from semantic entanglement and spatial biases. To fundamentally bypass this, we propose a purely spatially-agnostic feature representation using histograms.

Furthermore, to avoid the severe luminance-chrominance coupling inherent in the RGB space \cite{lvColorTransferImages2024}, we compute these statistical descriptors in the CIE Lab color space. Specifically, the input image is converted to the Lab space, and we extract three spatially-agnostic statistical descriptors: a 1D luminance histogram $H_L$ (capturing independent illumination), a 2D color histogram $H_{ab}$ (capturing independent chrominance), and a color-conditioned mean luminance map $M_{L|ab}$ (capturing luminance-chrominance correlation). The detailed architectural pipeline, mathematical formulations, and robust extraction strategies (e.g., non-linear stretching and soft-binning) of the proposed Lab-Extractor are provided in Appendix \ref{sec:appendix_extractor}.

\subsection{Attention-Guided Residual LUT Generation via MR-Mapper}
\label{sec:prediction}

\begin{figure}[htbp]
    \centering
    \includegraphics[width=0.8\linewidth]{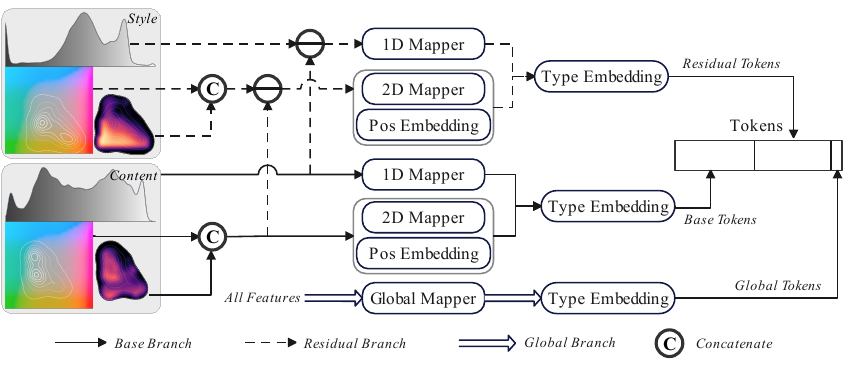}
    \caption{Architecture of the Multi-dimensional Residual Mapper (MR-Mapper), which unifies heterogeneous input features into a standard token sequence via Base, Residual, and Global branches.}
    \label{fig:mr_mapper}
\end{figure}

We formulate 3D LUT prediction as a Transformer-based Sequence-to-Sequence (Seq2Seq) translation task. To effectively inject the heterogeneous statistical features (1D/2D histograms and conditional luminance) into the network, we propose the Multi-dimensional Residual Mapper (MR-Mapper) to unify them into a standard token sequence (Figure \ref{fig:mr_mapper}). This holistic prediction ensures mapping smoothness. Beyond the inherent temporal consistency of global LUTs ($f(x)=f(x)$), such smoothness suppresses flickering caused by minor pixel fluctuations ($x_1 \approx x_2 \Rightarrow f(x_1) \approx f(x_2)$).

The MR-Mapper comprises three distinct branches. The \textit{Base Branch} encodes the content image's physical distribution, injecting 2D positional embeddings to preserve CIE Lab topological coordinates. The \textit{Residual Branch} computes statistical differences ($\ominus$ in Figure \ref{fig:mr_mapper}) between style and content features, explicitly guiding the network to learn color shifts. The \textit{Global Branch} extracts macroscopic statistics (e.g., contrast difference, chroma shift) as global constraints. Finally, branch outputs ($F_{base}$, $F_{res}$, $F_{global}$) are summed with type embeddings ($E_{type}$) and concatenated into a unified condition memory $M$:
\begin{equation}
    M = \text{Concat}(F_{base}, F_{res}, F_{global}) + E_{type}
\end{equation}

This memory $M$ serves as the Key ($K$) and Value ($V$) in a Transformer decoder, while a 3D Identity LUT grid ($LUT_{id} \in \mathbb{R}^{N \times 3}$, $N=D^3$) acts as the Query ($Q$). To preserve color lattice topology, learnable 3D positional encodings ($PE_R, PE_G, PE_B$) are added to $Q$. The queries, keys, and values are formulated as:
\begin{equation}
    Q = W_q(LUT_{id}) + (PE_R \oplus PE_G \oplus PE_B), \quad K = W_k(M), \quad V = W_v(M)
\end{equation}
where $W_q, W_k, W_v$ are linear projections, and $\oplus$ denotes broadcast addition.

Through cross-attention, each lattice point adaptively aggregates color transformation information to predict a color residual $\Delta C$ via a Feed-Forward Network (FFN):
\begin{equation}
    \Delta C = \text{FFN}\left( \text{Softmax}\left(\frac{Q K^T}{\sqrt{d_k}}\right) V \right)
\end{equation}
Predicting residuals aligns with the physical nature of color shifting and significantly eases optimization. The final stylized LUT is:
\begin{equation}
    LUT_{pred} = \text{Clamp}(LUT_{id} + \Delta C, 0, 1)
\end{equation}
To ensure training stability, we zero-initialize the final FFN projection layer. This guarantees an initial identity mapping ($\Delta C = 0$), preventing early-stage color distortion.

To eliminate the reliance on expensive paired datasets, we design an efficient self-supervised paradigm. We synthesize ground-truth stylized images $I_{gt}$ by applying random LUTs ($LUT_{gt}$) to content images $I_c$. Spatial augmentations (e.g., cropping, scaling) are then applied to $I_{gt}$ to generate the style reference $I_s$. Crucially, because our Lab-Extractor is strictly spatially-agnostic, it fundamentally prevents the network from taking a "shortcut" by memorizing spatial alignments, enabling robust training without semantic entanglement.

The training objective comprises three components. First, Smooth L1 loss constrains 3D color space:
\begin{equation}
    \mathcal{L}_{LUT} = \mathcal{L}_{\text{SmoothL1}}(LUT_{pred}, LUT_{gt})
\end{equation}
Second, we ensure image-domain fidelity via pixel-wise L1 and SSIM \cite{wangImageQualityAssessment2004} losses between the stylized output $I_{pred} = LUT_{pred}(I_c)$ and $I_{gt}$:
\begin{equation}
    \mathcal{L}_{img} = \| I_{pred} - I_{gt} \|_1 + \mathcal{L}_{SSIM}(I_{pred}, I_{gt})
\end{equation}
Finally, monotonicity ($\mathcal{L}_{mono}$) and total variation ($\mathcal{L}_{tv}$) regularization terms are applied to maintain a stable color manifold. The total loss is:
\begin{equation}
    \mathcal{L}_{total} = \lambda_{lut} \mathcal{L}_{LUT} + \lambda_{img} \mathcal{L}_{img} + \lambda_{mono} \mathcal{L}_{mono} + \lambda_{tv} \mathcal{L}_{tv}
\end{equation}

\subsection{Text-Driven Feature Synthesis via H-Diffuser}
\label{sec:text-driven}

To bypass the computationally expensive two-stage paradigm (text-to-image generation followed by style transfer) and the under-constrained nature of directly predicting 3D LUTs from text, we propose the H-Diffuser (Histogram Diffuser). It is a lightweight Diffusion Transformer (DiT) designed to directly synthesize spatially-agnostic statistical features from a text prompt $T_s$. Specifically, we flatten and concatenate the target features ($H_L$, $H_{ab}$, $M_{L|ab}$) into a 1D token sequence $X_0$.

Unlike standard diffusion models that predict added noise, H-Diffuser directly predicts the clean sequence $\hat{X}_0$ conditioned on CLIP\cite{radfordLearningTransferableVisual2021} text embeddings. This $X_0$-prediction objective is crucial for enforcing the strict physical boundaries (e.g., non-negativity) of statistical histograms. The predicted tokens are then linearly projected and reshaped back to their original spatial dimensions to reconstruct $\hat{H}_L$, $\hat{H}_{ab}$, and $\hat{M}_{L|ab}$.

A critical challenge in generating color histograms is the extreme sparsity of the 2D chrominance space, where most bins are empty. To prevent the network from predicting overly smoothed distributions dominated by these empty bins, we propose a density-aware loss. We construct a spatial weight mask $W = 1 + \alpha \bar{H}_{ab}^{gt}$, where $\bar{H}_{ab}^{gt} \in [0, 1]$ is the normalized ground-truth chrominance distribution and $\alpha$ is a scaling factor. This mask explicitly penalizes errors more heavily in dense color clusters. The final objective combines a Smooth L1 loss for the 1D luminance and a mask-weighted L1 loss for the sparse 2D sequences:
\begin{equation}
    \mathcal{L}_{diff} = \lambda_{L} \mathcal{L}_{\text{SmoothL1}}(\hat{H}_{L}, H_{L}^{gt}) + \big\| W \odot \big( \lambda_{ab}(\hat{H}_{ab} - H_{ab}^{gt}) + \lambda_{abL}(\hat{M}_{L|ab} - M_{L|ab}^{gt}) \big) \big\|_1
\end{equation}

where $\odot$ denotes element-wise multiplication. Once generated, these text-driven statistical features seamlessly replace the reference-extracted features to generate the stylized 3D LUT.

\section{Experiments}
\label{sec:exp}

We evaluate StatLUT against state-of-the-art methods. We detail the experimental setup, followed by qualitative and quantitative comparisons (Sec. \ref{sec:exp-com}), a user study (Sec. \ref{sec:exp-user-study}), and ablation studies (Sec. \ref{sec:ablation}). Implementation details are provided in Appendix \ref{sec:appendix_implementation}. Video color grading results demonstrating temporal consistency are provided in Appendix \ref{sec:appendix_video}. Experimental results for cross-semantic scenarios are provided in Appendix \ref{sec:appendix_cross_semantic}.

\textbf{Datasets.} We train on MS COCO \cite{linMicrosoftCOCOCommon2014} for content images and a custom dataset of 10,000 diverse 3D LUTs for self-supervised style conditions. Evaluation is conducted on the PST50(CC-BY 4.0 license) \cite{gongSALUTSpatialAdaptive2025} and PhotoNAS(MIT license) \cite{anUltrafastPhotorealisticStyle2020} benchmarks, alongside high-resolution images for generalization assessment.

\textbf{Implementation Summary.} Images are resized to $224 \times 224$. The Lab-Extractor quantizes the color space to yield a 2304-dimensional statistical feature, which modulates a $16^3$ 3D LUT via a Transformer decoder. For text-driven generation, we employ a frozen CLIP encoder and an 8-block DiT (H-Diffuser) sampled via 200-step DDIM \cite{songDenoisingDiffusionImplicit2021}.

\textbf{Quantitative Metrics.} Since traditional metrics (e.g., PSNR, LPIPS) often misalign with human perception in photorealistic style transfer, we adopt the specialized protocol from Neural Preset \cite{keNeuralPresetColor2023}. We report style similarity (via a pre-trained discriminator) and content similarity (SSIM on LDC \cite{soriaLdcLightweightDense2022} features), both normalized to $[0, 1]$. Efficiency is measured by learnable parameters (Params) and multiply-accumulate operations (MACs).

\subsection{Comparisons}
\label{sec:exp-com}
We compare our proposed StatLUT against state-of-the-art deep learning-based color mapping methods: NLUT\cite{chenNLUTNeuralbased3D2023}, Neural Preset\cite{keNeuralPresetColor2023}, D-LUT\cite{liDLUTPhotorealisticStyle2025}, SA-LUT\cite{gongSALUTSpatialAdaptive2025}, and DLUT-VCG\cite{shinVideoColorGrading2025}. These methods serve as highly relevant baselines since they all generate explicit parameters for color mapping. For the NAS dataset, we additionally include the PhotoNAS\cite{anUltrafastPhotorealisticStyle2020} results provided by the original authors. To ensure a fair comparison, all baselines are evaluated using their official pre-trained models or default configurations. Quantitative results are computed comprehensively across the entire set of 90 content-style pairs to guarantee an unbiased evaluation.

\subsubsection{Qualitative Results}

\begin{figure}[htbp]
    \centering
    \includegraphics[width=1\linewidth]{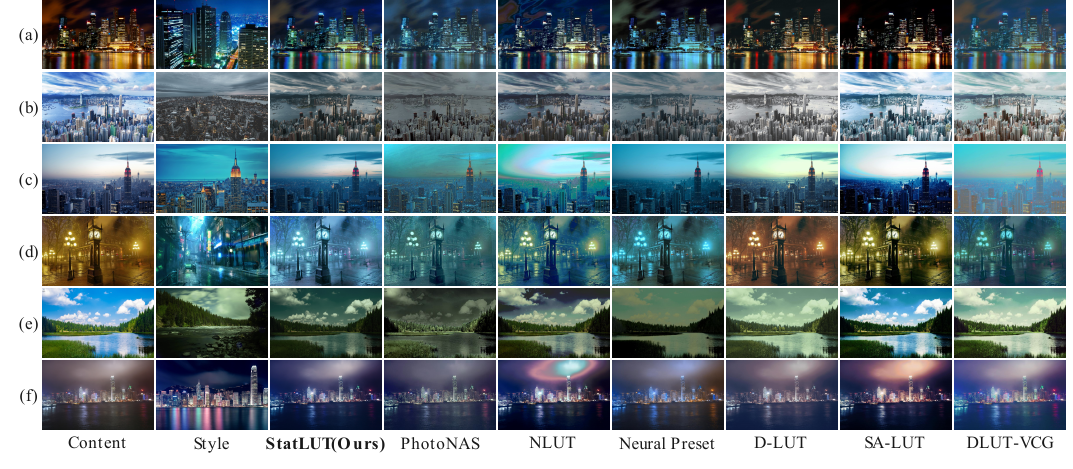}
    \caption{Visual evaluation of different methods on the NAS dataset.}
    \label{fig:nas_compares}
\end{figure}

\begin{figure}[htbp]
    \centering
    \includegraphics[width=1\linewidth]{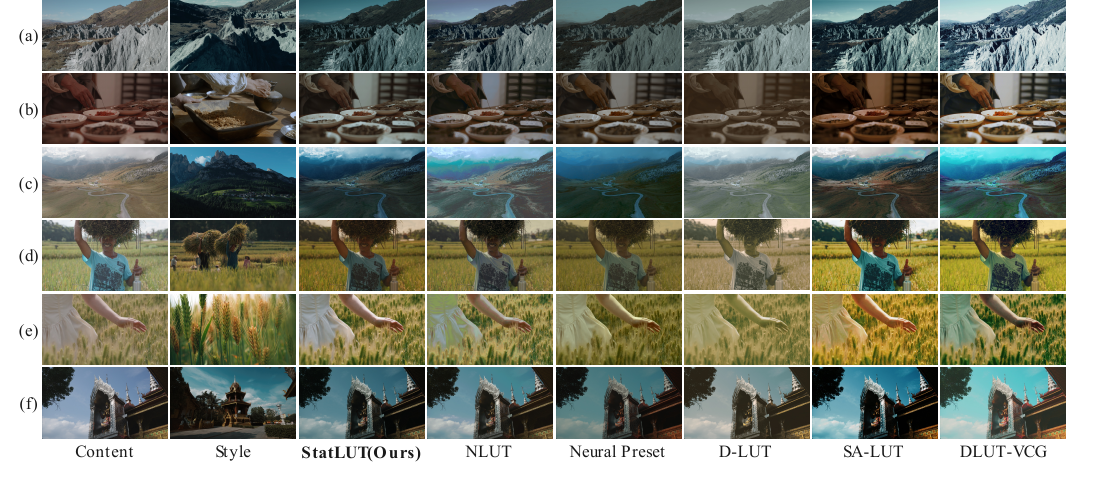}
    \caption{Visual evaluation of different methods on the PST50 dataset.}
    \label{fig:pst_compares}
\end{figure}

Figure \ref{fig:nas_compares} presents a visual comparison on the NAS dataset. Overall, StatLUT successfully achieves a color distribution that closely matches the reference style while faithfully preserving the structural integrity of the content images. In contrast, baseline methods exhibit various visual artifacts. Specifically, PhotoNAS and DLUT-VCG suffer from severe contrast degradation in certain scenes. NLUT, while effective in some cases, introduces pronounced color banding artifacts, particularly in smooth regions like skies. Neural Preset struggles with insufficient contrast, leading to blurred structural boundaries. Furthermore, D-LUT and SA-LUT demonstrate limited stylization capacity, resulting in outputs that noticeably deviate from the target styles.

Figure \ref{fig:pst_compares} illustrates the results on the PST50 dataset, where StatLUT consistently delivers superior and natural color transfer. Among the baselines, NLUT continues to exhibit color banding, whereas Neural Preset yields slightly degraded visual quality due to diminished global contrast. D-LUT shows inadequate stylization and occasionally introduces erroneous color mappings, such as unnatural skin tone renditions. SA-LUT tends to produce overly saturated results that fail to align with the style references. Lastly, DLUT-VCG demonstrates inconsistent performance; although it achieves reasonable local stylization, it is prone to unnatural global color shifts.

Generally, the NAS dataset presents a greater challenge due to larger domain gaps between content and style images, which exposes the limitations of the baseline methods. While the PST50 dataset is less challenging, allowing most methods to achieve basic color transfer, the baselines still suffer from subtle detail degradation. Across both datasets, StatLUT significantly outperforms competing methods by striking an optimal balance between accurate style rendition and robust content texture preservation.

\subsubsection{Quantitative Results}

\begin{table*}[htbp]
    \caption{Quantitative comparisons on the NAS and PST50 datasets. StatLUT (Img) achieves the shortest distance to the ideal point, indicating an optimal balance between content preservation and style transfer. StatLUT (Text) relies solely on textual prompts without reference images. Best and second-best image-driven results are \textbf{bolded} and \underline{underlined}, respectively.}
    \label{tab:quantitative_results}
    \centering
    \resizebox{\textwidth}{!}{
        \begin{tabular}{llcccccccc}
            \toprule
            Dataset & Metric                      & PhotoNAS & NLUT              & Neural Preset  & D-LUT & SA-LUT & DLUT-VCG & StatLUT (Img)     & StatLUT (Text) \\
            \midrule
            NAS     & Content Sim. $\uparrow$     & 0.477    & 0.640             & \textbf{0.709} & 0.700 & 0.696  & 0.629    & \underline{0.707} & 0.688          \\
                    & Style Sim. $\uparrow$       & 0.557    & \textbf{0.787}    & 0.656          & 0.562 & 0.426  & 0.676    & \underline{0.746} & 0.498          \\
                    & Dist. to Ideal $\downarrow$ & 0.681    & \underline{0.400} & 0.448          & 0.517 & 0.598  & 0.491    & \textbf{0.386}    & 0.565          \\
            \midrule
            PST50   & Content Sim. $\uparrow$     & ---      & 0.746             & \textbf{0.774} & 0.734 & 0.729  & 0.713    & \underline{0.760} & 0.757          \\
                    & Style Sim. $\uparrow$       & ---      & \underline{0.849} & 0.738          & 0.467 & 0.650  & 0.549    & \textbf{0.870}    & 0.536          \\
                    & Dist. to Ideal $\downarrow$ & ---      & \underline{0.284} & 0.345          & 0.544 & 0.438  & 0.514    & \textbf{0.259}    & 0.487          \\
            \bottomrule
        \end{tabular}}
\end{table*}

\begin{figure}[htbp]
    \centering
    \includegraphics[width=1\linewidth]{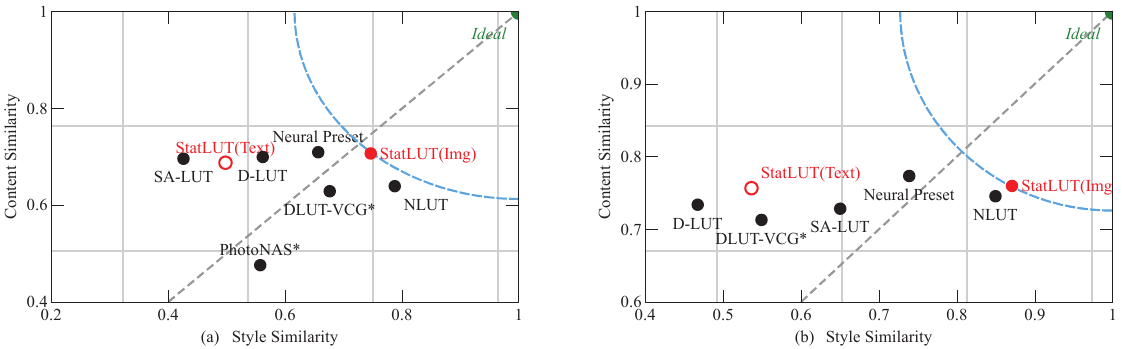}
    \caption{Quantitative evaluation on the (a) NAS and (b) PST50 datasets. The top-right corner $(1, 1)$ represents the ideal point. StatLUT (Img) (solid red dot) achieves the best balance between content and style similarity. Methods marked with (*) were resized during evaluation due to resolution mismatches. The hollow red circle denotes our text-driven variant, StatLUT (Text).}
    \label{fig:nas_pst_results}
\end{figure}

Figure \ref{fig:nas_pst_results} presents the quantitative evaluation. As depicted, StatLUT (Img) consistently positions closest to the ideal top-right corner across both datasets, demonstrating an optimal balance between structural preservation and style transfer. While NLUT achieves comparable style similarity, it suffers a notable drop in content similarity (e.g., $9.48\%$ lower than StatLUT on the NAS dataset). This degradation directly corroborates the color banding artifacts observed in Figure \ref{fig:nas_compares}, which inevitably corrupt underlying image textures. Conversely, Neural Preset maintains high content similarity but struggles with style transfer ($15.17\%$ lower style similarity than StatLUT on the PST50 dataset), aligning with its limited stylization capacity shown in Figure \ref{fig:pst_compares}.

\subsubsection{Cross-modal Text-driven Performance}
\label{sec:exp-text}

\begin{figure*}[htbp]
    \centering
    \includegraphics[width=\textwidth]{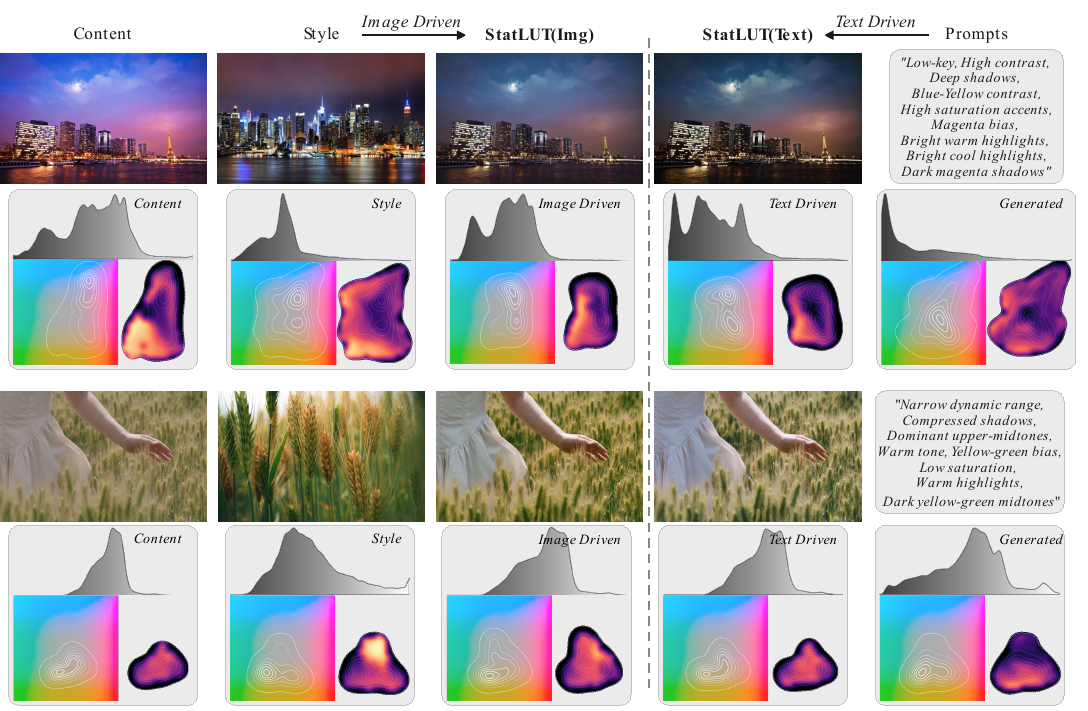}
    \caption{Visual comparison of intermediate features and stylized results between the image-driven (left) and text-driven (right) modes of StatLUT. Visualized features include 1D luminance histograms (top), 2D joint color distributions (bottom left), and mean luminance conditioned on color bins (bottom right).}
    \label{fig:multi_pst}
\end{figure*}

Existing text-driven editing methods predominantly focus on localized manipulations rather than global color mapping. Due to this task disparity, direct comparisons are infeasible. Therefore, we evaluate our cross-modal variant, StatLUT (Text), against our image-driven model as an upper-bound oracle.

Qualitatively (Figure \ref{fig:multi_pst}), despite relying solely on tonal descriptions without visual references, StatLUT (Text) generates intermediate statistical features (Column 5) that closely align with those extracted from actual style images (Column 2). Consequently, the text-driven outputs (Column 4) visually approximate the image-driven results (Column 3), demonstrating highly effective cross-modal feature alignment.

Quantitatively (Table \ref{tab:quantitative_results} and Figure \ref{fig:nas_pst_results}), StatLUT (Text) maintains highly competitive Content Similarity, ensuring robust structural preservation. Although its Style Similarity naturally trails the image-driven oracle due to the inherent cross-modal gap between abstract text and concrete visual references, StatLUT (Text) offers precise and controllable global color grading without compromising structural integrity.

\subsubsection{Semantic Decoupling Analysis}
\label{sec:exp-decoupling}

As emphasized in our motivation, existing methods often suffer from semantic entanglement, where the spatial structure of the style image inadvertently interferes with the color mapping process. To rigorously validate that StatLUT successfully decouples spatial semantics from color and luminance information, we design a novel style patch-shuffling experiment.

Specifically, we divide the style image into $P \times P$ patches and randomly shuffle them, which completely destroys its structural semantics while perfectly preserving its global color distributions. When applying these shuffled style images to the same content image, StatLUT maintains highly consistent color mapping results with near-zero residuals, significantly outperforming all baseline methods. This confirms that our statistical feature-driven paradigm is fundamentally immune to spatial semantic interference. We strongly encourage readers to refer to Appendix \ref{sec:appendix_patch_shuffle} for detailed visual and quantitative analyses of this compelling experiment.

\subsection{User Study}
\label{sec:exp-user-study}

Since quantitative metrics may not perfectly align with human perception, we conducted a user study with 50 participants (yielding 1,000 valid votes; see \ref{sec:user_details} for detailed protocols). As reported in Table \ref{tab:user_study}, our StatLUT significantly outperforms all baselines, achieving a dominant Top-1 preference rate of 70\% and the best average rank of 1.40 (compared to the runner-up Neural Preset at $15\%$ and $2.75$). This subjective evaluation strongly corroborates our quantitative findings, demonstrating that StatLUT consistently produces the most visually pleasing results.

\begin{table*}[htbp]
    \centering
    \caption{Quantitative results of the user study. We report the Top-1 preference rate ($\uparrow$) and the average rank ($\downarrow$) across 1,000 valid votes. The best results are highlighted in \textbf{bold}.}
    \label{tab:user_study}
    \resizebox{\textwidth}{!}{
        \begin{tabular}{lcccccc}
            \toprule
            Metric                           & NLUT & Neural Preset & D-LUT & SA-LUT & DLUT-VCG & StatLUT (Ours) \\
            \midrule
            Top-1 Preference (\%) $\uparrow$ & 10   & 15            & 0     & 5      & 0        & \textbf{70}    \\
            Average Rank $\downarrow$        & 2.95 & 2.75          & 4.70  & 4.95   & 4.25     & \textbf{1.40}  \\
            \bottomrule
        \end{tabular}
    }
\end{table*}

\subsection{Ablation Study}
\label{sec:ablation}

To validate our key components, we conduct ablation studies under identical settings.

\begin{table}[htbp]
    \centering
    \small
    \begin{minipage}[t]{0.48\textwidth}
        \centering
        \caption{Ablation on feature combinations.}
        \label{tab:feature_ablation}
        \vspace{-8pt}
        \begin{tabular}[t]{l|ccc}
            \toprule
            Features        & Cont. $\uparrow$ & Style $\uparrow$ & Dist. $\downarrow$ \\
            \midrule
            A: RGB 1D-hist  & 0.714            & \textbf{0.874}   & 0.286              \\
            B: Lab L+ab     & 0.716            & 0.848            & 0.304              \\
            C: Lab L+ab+abL & \textbf{0.726}   & 0.871            & \textbf{0.281}     \\
            \bottomrule
        \end{tabular}
    \end{minipage}
    \hfill
    \begin{minipage}[t]{0.48\textwidth}
        \centering
        \caption{Ablation on mapping modules.}
        \label{tab:mapper_ablation}
        \vspace{-8pt}
        \begin{tabular}[t]{l|cc|c}
            \toprule
            Mapper             & Cont. $\uparrow$ & Dist. $\downarrow$ & Params $\downarrow$ \\
            \midrule
            Simple MLP         & 0.723            & 0.286              & 428.9M              \\
            Bottleneck         & 0.726            & 0.281              & 5.24M               \\
            \textbf{MR-Mapper} & \textbf{0.760}   & \textbf{0.259}     & \textbf{0.38M}      \\
            \bottomrule
        \end{tabular}
    \end{minipage}
\end{table}

\textbf{Statistical Feature Combinations.} As shown in Table \ref{tab:feature_ablation}, compared to highly coupled RGB histograms (Features A) or purely disentangled Lab histograms (Features B), our complete set (Features C) explicitly re-establishes the binding relationship between specific colors and illumination. It provides the most precise prior, achieving the optimal Dist. to Ideal ($0.281$).

\textbf{Condition Mapping Module.} Table \ref{tab:mapper_ablation} compares our MR-Mapper against MLP baselines using Features C. Traditional MLPs destroy structural information or introduce massive redundancy. In contrast, our MR-Mapper explicitly models residual differences and global statistics, achieving the highest generation quality (Content Sim. $0.760$) with a compact footprint of $0.38$M parameters.

\textbf{LUT Size Selection.} We also investigate the 3D LUT size $D$. Considering the cubic spatial complexity $\mathcal{O}(D^3 \times 3)$, we observe diminishing marginal utility beyond $D=16$. Thus, we set $D=16$ to balance fidelity and efficiency (see Appendix \ref{sec:ablation_details}).

\section{Limitations and Future Work}
\label{sec:limitations}

Despite achieving state-of-the-art performance, StatLUT has limitations. First, our pipeline relies on absolute style descriptions. Future work could adapt H-Diffuser for relative textual prompts (e.g., ``make it warmer'') for intuitive editing. Second, the global nature of 3D LUTs precludes semantic-aware local fine-tuning. Exploring 4D LUTs with a semantic channel could address this while maintaining efficiency. Finally, extreme content-style gaps (e.g., daytime snow to nighttime neon) may yield suboptimal results. Integrating traditional statistical matching as a fallback for such outliers remains a practical necessity.

\section{Conclusion}
\label{sec:conclusion}

In this paper, we present StatLUT, a novel multimodal framework that advances photorealistic style transfer (PST) via statistical feature-driven 3D LUT generation. To resolve semantic entanglement and spatial leakage, we extract spatially-agnostic statistical distributions in the CIE Lab space as explicit color priors, ensuring strict structural preservation. By formulating LUT prediction as a Transformer-based Seq2Seq task with our MR-Mapper, StatLUT globally models color topology with a highly compact parameter footprint, effectively eliminating color banding. Furthermore, we pioneer text-driven PST via the H-Diffuser, which seamlessly synthesizes statistical features from tone-descriptive prompts. Extensive experiments confirm that StatLUT achieves state-of-the-art performance in color fidelity, structural integrity, and multimodal flexibility. We envision this statistical prior paradigm as a highly promising direction for high-resolution image and video color grading.

\begin{ack}
    This work was supported by Honor Device Co., Ltd. The authors declare no competing interests.
\end{ack}

\small

% 设置参考文献样式为 unsrtnat（按正文引用顺序编号的数字格式）
\bibliographystyle{unsrtnat}

% 引入您的 .bib 文件（假设您的文件名为 references.bib）
\bibliography{StatLUT-Ref}

\clearpage
\appendix
\section{Detailed Formulation of the Lab-Extractor}
\label{sec:appendix_extractor}

\begin{figure}[htbp]
    \centering
    \includegraphics[width=1\linewidth]{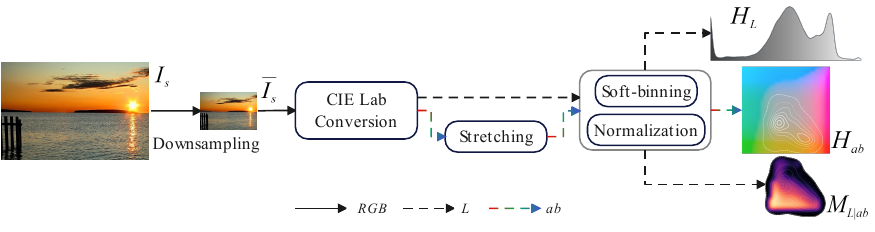}
    \caption{Pipeline of the proposed Lab-Extractor. The downsampled image is converted to the CIE Lab space to decouple luminance and chrominance. Following non-linear stretching of the $a$ and $b$ channels, three spatially-agnostic statistical descriptors ($H_L$, $H_{ab}$, and $M_{L|ab}$) are extracted via soft-binning. Contour lines in $H_{ab}$ denote color frequency, where innermost regions indicate dominant colors. Overlaying these contours on $M_{L|ab}$ visually highlights the luminance properties corresponding to these dense color regions.}
    \label{fig:lab-extractor}
\end{figure}

As discussed in Section \ref{sec:decoupling}, the input image is first spatially downsampled and converted to the Lab space to ensure efficient computation and strict decoupling of luminance and chrominance. In natural images, chrominance values ($a$ and $b$) typically concentrate heavily around zero (neutral grays), causing sparse activations in peripheral histogram bins. To address this long-tail distribution and maximize representational capacity, we apply a \textbf{non-linear stretching} to the chrominance channels before quantization:
$$
    \hat{c} = \text{sgn}(c) \cdot \left( \frac{|c|}{S} \right)^\gamma
$$
where $c \in \{a, b\}$, $\hat{c}$ is the stretched output, $\text{sgn}(\cdot)$ is the sign function, and $S$ is a normalization scaling factor. The hyperparameter $\gamma < 1$ expands the central dense regions while compressing the sparse extremes, yielding a more uniform and informative distribution across color bins.

As illustrated in Figure \ref{fig:lab-extractor}, the Lab-Extractor derives three statistical components from the stretched Lab space: $H_L$, $H_{ab}$, and $M_{L|ab}$. The discrete extraction is formulated as:
$$
    \begin{aligned}
        H_L(i)        & = \frac{1}{HW} \sum_{x=1}^{W} \sum_{y=1}^{H} \mathbb{I}\big(L(x,y) \in \mathcal{B}_i^L\big)                                                                                                                                                                                           \\
        H_{ab}(j,k)   & = \frac{1}{HW} \sum_{x=1}^{W} \sum_{y=1}^{H} \mathbb{I}\big(\hat{a}(x,y) \in \mathcal{B}_j^a \land \hat{b}(x,y) \in \mathcal{B}_k^b\big)                                                                                                                                              \\
        M_{L|ab}(j,k) & = \frac{\sum_{x=1}^{W} \sum_{y=1}^{H} L(x,y) \cdot \mathbb{I}\big(\hat{a}(x,y) \in \mathcal{B}_j^a \land \hat{b}(x,y) \in \mathcal{B}_k^b\big)}{\sum_{x=1}^{W} \sum_{y=1}^{H} \mathbb{I}\big(\hat{a}(x,y) \in \mathcal{B}_j^a \land \hat{b}(x,y) \in \mathcal{B}_k^b\big) + \epsilon}
    \end{aligned}
$$
where $H$ and $W$ are the image dimensions, $(x,y)$ are spatial coordinates, and $\mathbb{I}(\cdot)$ is the indicator function. $\mathcal{B}_i^L$, $\mathcal{B}_j^a$, and $\mathcal{B}_k^b$ denote the $i$-th, $j$-th, and $k$-th bins for the $L$, $\hat{a}$, and $\hat{b}$ channels, respectively, and $\epsilon$ prevents division by zero.

In practice, to avoid quantization artifacts from the hard indicator $\mathbb{I}(\cdot)$, we employ a \textbf{soft-binning strategy}. Specifically, bilinear interpolation is used to distribute each pixel's contribution across adjacent bins, yielding a smoother and more robust descriptor. Finally, to prevent dominant colors (e.g., large background areas) from overwhelming the histogram, we apply a \textbf{square-root normalization} to the 2D color histogram: $\hat{H}_{ab}(j,k) = \sqrt{H_{ab}(j,k) + \epsilon}$. This non-linear dampening balances feature magnitudes, ensuring minority colors still contribute effectively to the style representation.

\section{Detailed Experimental Setups}
\label{sec:appendix_implementation}

\textbf{Detailed Datasets.}
To establish style conditions for our self-supervised paradigm, our constructed dataset comprises 10,000 3D LUTs, where 4,000 are sourced from professional color grading resources and 6,000 are synthesized via statistical distribution sampling to guarantee stylistic diversity.

\textbf{Image-Driven Pipeline Details.}
During training, the proposed Lab-Extractor quantizes the luminance $H_L$ into 256 bins, and both $H_{ab}$ and $M_{L|ab}$ into $32 \times 32$ grids, resulting in the 2304-dimensional feature. For chrominance stretching, we set $\gamma = 0.5$ and a scale factor of $S = 128$. The Transformer decoder features 6 layers with a hidden dimension of $d_{model} = 512$ and 8 attention heads. The network is optimized using AdamW \cite{loshchilovDecoupledWeightDecay2019} for 200 epochs. The learning rate is initialized at $3.0 \times 10^{-4}$ with a weight decay of 0.05, incorporating a 5-epoch linear warmup followed by a cosine annealing schedule decaying to $1.0 \times 10^{-7}$. The empirical weights for the objective functions are set to $\lambda_{lut} = 1.0$, $\lambda_{img} = 0.5$, $\lambda_{mono} = 5.0$, and $\lambda_{tv} = 0.0001$. The model is trained on 8 NVIDIA A800 GPUs for approximately 15 hours. During inference on a single A800 GPU, the entire image-driven pipeline (including feature extraction, LUT prediction, and mapping) takes less than 50 ms per image. Notably, the final 3D LUT mapping step is extremely efficient, requiring less than 0.1 ms.

\textbf{Text-Driven Pipeline Details.}
To construct the training pairs, we employ Qwen3.5\cite{qwen3.5} to generate stylistic captions for all images in the COCO dataset. The generation process is strictly constrained to describe only luminance, chrominance, and their coupling effects, explicitly excluding any object-level semantics. These captions are then paired with the corresponding 2304-dimensional statistical features extracted by our Lab-Extractor to form the text-feature dataset. For the H-Diffuser, the 2304-dimensional features are tokenized into 36 patches, processed by an 8-block DiT architecture (hidden size of 768 and 12 heads). The diffusion model is trained for 500K iterations on 8 NVIDIA A800 GPUs with a total batch size of 4096. We apply a learning rate of $1.0 \times 10^{-4}$ with a 3000-step linear warmup and a cosine decay schedule. The balancing weights for the diffusion objective are empirically set to $\lambda_L = 1.0$, $\lambda_{ab} = 1.2$, and $\lambda_M = 1.5$. The forward diffusion process spans 1000 timesteps. The entire training process takes approximately 50 hours. During inference on a single A800 GPU, the text-driven pipeline takes about 2 seconds per image, with the final 3D LUT mapping step remaining highly efficient at less than 0.1 ms. The substantial latency of the text-driven branch originates from the iterative generation process from text to features, which can be formulated as $\text{MACs}_{\mathit{total}} = \text{MACs}_{\mathit{CLIP}} + 200 \times \text{MACs}_{\mathit{DiT}_{\mathit{step}}}$ due to the 200-step DDIM sampling. However, this process is only triggered when the text description is updated. In most practical applications, the inference time is close to that of the image-driven branch (less than 50 ms).

\section{Application to Video Color Grading}
\label{sec:appendix_video}
To further demonstrate the practical applicability and robustness of our proposed StatLUT, we apply it to video color grading. In this experiment, we extract the color transformation from a single reference style image and apply it frame-by-frame to a source video.

\begin{figure}[htbp]
    \centering
    \includegraphics[width=1\linewidth]{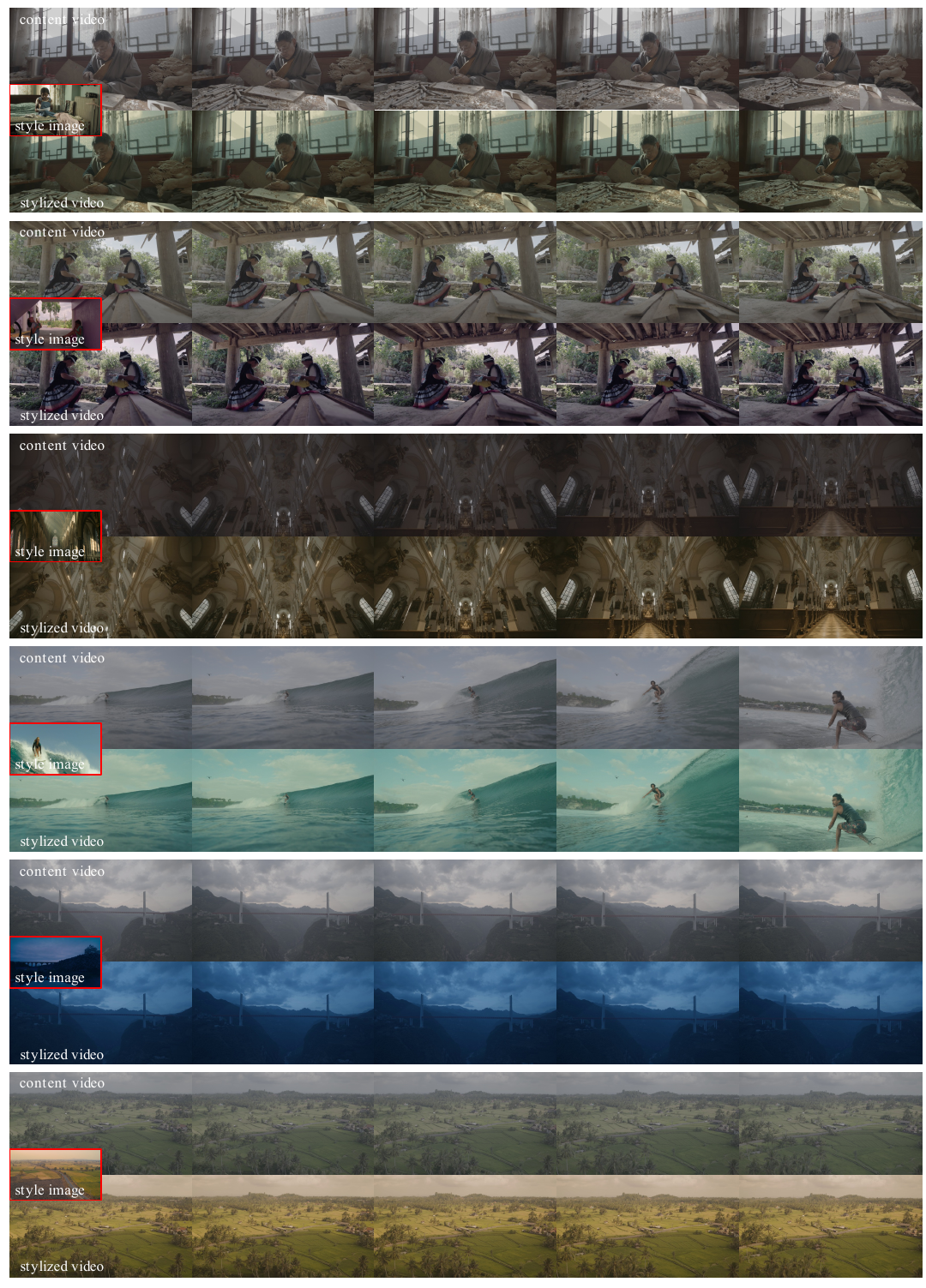}
    \caption{Qualitative results of video color grading. For each example, the top row shows the original content video frames, the left inset is the reference style image, and the bottom row presents our stylized video frames. Our method effectively transfers the desired color style while maintaining temporal consistency across consecutive frames.}

    \label{fig:video_results}
\end{figure}

As shown in Figure \ref{fig:video_results}, our method successfully transfers the desired color style to the target video sequence. More importantly, because our method relies on stable statistical color mapping, the stylized video maintains excellent temporal consistency without introducing noticeable flickering or artifacts across consecutive frames. This indicates that StatLUT can be naturally extended to video processing tasks.

\section{Generalization to Cross-Semantic Scenarios}
\label{sec:appendix_cross_semantic}
\textbf{Experimental Setup.}
To further demonstrate the generalization capability of StatLUT for PST tasks across diverse semantic scenarios, we selected style images from the WikiArt\cite{PhillipsWiki2011} dataset, covering various artistic movements such as Abstract Expressionism, Impressionism, and Rococo. These styles present a significant domain gap compared to the PST50 dataset. To maximize the semantic challenge, we deliberately paired content and style images from entirely different categories, including portraits, architecture, landscapes, and objects.

\begin{figure}[htbp]
    \centering
    \includegraphics[width=1\linewidth]{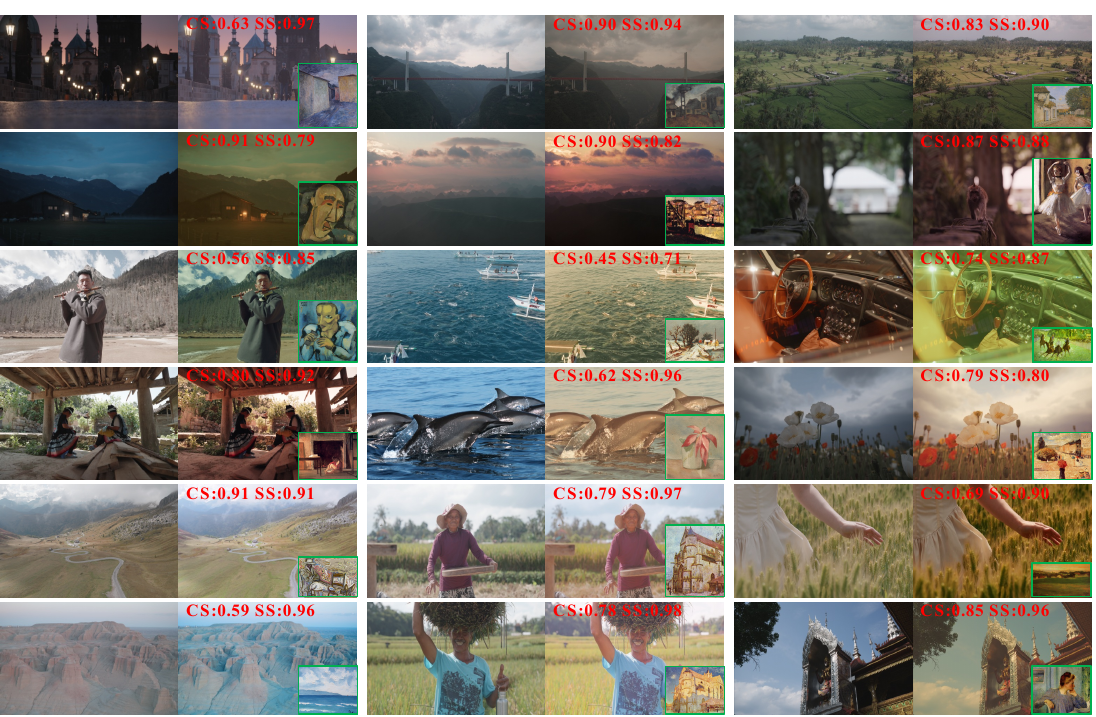}
    \caption{Cross-semantic results (PST50 content, WikiArt style). For each pair: content (left), stylized result (right), and style reference (inset). CS/SS denote content/style similarity.}

    \label{fig:cross_semantic}
\end{figure}

\textbf{Results and Analysis.}
As illustrated in Figure \ref{fig:cross_semantic}, we present PST results across 18 diverse cross-semantic pairs. The consistently high similarity scores suggest that the performance of StatLUT is not strictly dependent on semantic alignment between the content and style images. A representative case is the transfer of a classical portrait style to ancient temple architecture (Row 6, Col 3), which achieves a Style Similarity (SS) of 0.96 and a Content Similarity (CS) of 0.85. Such stability, even amidst significant semantic gaps, validates our core design: by leveraging spatially-agnostic features to decouple structural semantics from luminance and color information, StatLUT ensures robust and natural color mapping even across disparate semantic domains.

\section{Semantic Decoupling Analysis: Patch-Shuffling Experiment}
\label{sec:appendix_patch_shuffle}

\begin{figure}[htbp]
    \centering
    \includegraphics[width=1\linewidth]{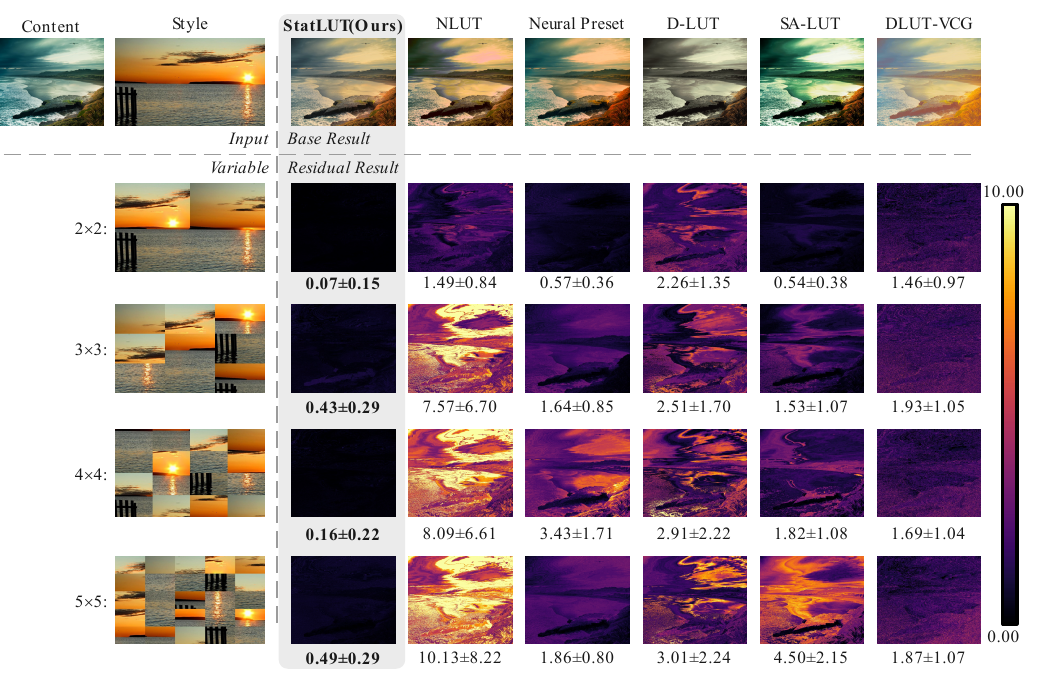}
    \caption{Visual and quantitative results of the style patch-shuffling experiment. The top row displays the content image, the original intact style image, and the base stylized results generated by different methods. The subsequent rows present the randomly shuffled style images ($P \times P$ patches, where $P \in \{2, 3, 4, 5\}$) alongside the corresponding residual maps. These maps represent the absolute difference between the results guided by the shuffled styles and the base results. The mean and standard deviation of the residuals are reported below each map. StatLUT (Ours) exhibits near-zero residuals (darkest maps), demonstrating superior robustness against spatial semantic interference.}
    \label{fig:patch_shuffle_appendix}
\end{figure}

In Section \ref{sec:exp-decoupling} of the main manuscript, we briefly introduced the style patch-shuffling experiment. Here, we provide the detailed experimental setups, visual comparisons, and quantitative analyses to further demonstrate the robust semantic decoupling capability of StatLUT.

\textbf{Experimental Setup.}
To explicitly isolate the influence of spatial structural semantics from color and luminance information, we divide a given reference style image into $P \times P$ patches (where $P \in \{2, 3, 4, 5\}$) and randomly shuffle their spatial positions. This operation completely disrupts the structural semantics (e.g., the shape of the sky, ocean, or sun) while strictly preserving the complete color and luminance distribution of the original style image. Subsequently, we perform style transfer on the same content image using these shuffled style variants. To quantify the stability of the color mapping, we calculate the residual maps between the stylized results guided by the shuffled styles and the result guided by the original, intact style image. We also record the mean and standard deviation of these residuals.

\textbf{Results and Analysis.}
As illustrated in Figure \ref{fig:patch_shuffle_appendix}, baseline methods (such as NLUT, D-LUT, and SA-LUT) exhibit significant variations and high residuals when the style image's spatial structure is altered. This clearly indicates that their color mapping processes are heavily entangled with the spatial semantics of the reference image.

In stark contrast, StatLUT consistently achieves the lowest residuals across all $P \times P$ configurations. The residual maps for our method are nearly entirely dark, proving that our Lab-Extractor effectively extracts pure statistical color features, rendering the color mapping process completely invariant to spatial structural changes.

It is worth noting that the residuals for StatLUT are not strictly zero (e.g., $0.07 \pm 0.15$ for the $2 \times 2$ configuration). This minor deviation is mathematically expected and reasonable. Before extracting the 1D and 2D histogram features, the input images undergo a resizing operation. For the shuffled style images, this resizing triggers resampling and interpolation at the newly formed artificial patch boundaries, causing slight pixel-level deviations from the original intact image. Consequently, these interpolation differences introduce negligible variations in the extracted statistical features, leading to the minimal non-zero residuals observed.

\section{User Study Details}
\label{sec:user_details}

To evaluate the perceptual quality and aesthetic appeal of our method in real-world scenarios, we conducted a comprehensive user study. We recruited 50 participants for this study. From the combined NAS and PST50 datasets (90 pairs in total), we randomly sampled 20 content-style pairs for each participant to mitigate visual fatigue.

For each trial, the participant was presented with the reference content and style images, followed by six stylized results generated by NLUT, Neural Preset, D-LUT, SA-LUT, DLUT-VCG, and our StatLUT. To ensure a fair comparison and eliminate positional bias, the six generated images were displayed in a completely randomized order for each trial. Participants were asked to rank the six results from 1 (best) to 6 (worst) based on three primary criteria: style similarity to the reference, content preservation (e.g., absence of artifacts), and overall visual aesthetics.

\section{Ablation Study Details}
\label{sec:ablation_details}

To validate the effectiveness of the key components proposed in our method, we conduct extensive ablation studies under identical datasets and training settings. This section primarily investigates the impact of the statistical feature combinations, the design of the condition mapping module, and the LUT size on the final color transfer performance and model efficiency.

\subsection{Effectiveness of Statistical Feature Combinations}
To evaluate the superiority of the statistical features designed in the Lab color space, we design three feature combinations for comparison. Since the feature dimensions vary across these variants, they cannot be directly processed by our proposed MR-Mapper. For a fair comparison, all three variants utilize a unified Bottleneck MLP combined with base token mapping to generate the condition vectors.

\begin{itemize}
    \item \textbf{Features A (RGB 1D-hist)}: Extracts independent 1D histograms of the three channels in the RGB space (yielding a $3 \times 256$-dimensional feature).
    \item \textbf{Features B (Lab L+ab)}: Extracts independent lightness histograms (L-hist, 256 bins) and 2D color histograms (ab-hist, $32 \times 32$ bins) in the Lab space.
    \item \textbf{Features C (Lab L+ab+abL)}: Employs the complete feature input proposed in this paper, adding the ab-conditioned mean lightness ($32 \times 32$ bins) to Features B.
\end{itemize}

The quantitative comparison results are presented in Table \ref{tab:feature_ablation_d}. By independently calculating the histograms in the RGB space (Features A), where the lightness and color information are highly coupled across the three channels, the model learns certain mapping relationships but achieves only suboptimal performance. When shifting to the Lab space (Features B) and completely disentangling lightness from color, providing only independent lightness and color histograms leads to the loss of correspondence between specific colors and their specific lightness levels (e.g., the blue of the sky is bright, while the green of the trees is dark). Consequently, the model becomes confused during color mapping, resulting in the lowest overall performance. Features C, by introducing the ab-conditioned mean lightness (abL), explicitly re-establishes the binding relationship between specific colors and illumination while maintaining the advantages of lightness-color decoupling. This design avoids the dimensional explosion associated with 3D joint histograms and provides the most precise color-lightness mapping prior, thereby achieving the optimal performance across all metrics (with the Dist. to Ideal dropping to $0.281$).

\begin{table}[htbp]
    \centering
    \caption{Ablation study on the input statistical feature combinations. All variants use a unified Bottleneck MLP for fair comparison.}
    \label{tab:feature_ablation_d}
    \resizebox{\textwidth}{!}{
        \begin{tabular}{l|ccc}
            \toprule
            Feature Components                          & Content Sim. $\uparrow$ & Style Sim. $\uparrow$ & Dist. to Ideal $\downarrow$ \\
            \midrule
            Features A: RGB 1D-hist                     & 0.714                   & \textbf{0.874}        & 0.286                       \\
            Features B: Lab L-hist + ab-hist            & 0.716                   & 0.848                 & 0.304                       \\
            Features C: Lab L-hist + ab-hist + abL-mean & \textbf{0.726}          & 0.871                 & \textbf{0.281}              \\
            \bottomrule
        \end{tabular}
    }
\end{table}

\subsection{Design of the Condition Mapping Module}
We further investigate the advantages of the proposed MR-Mapper in projecting statistical features into Transformer condition vectors by comparing it against two baseline mapping strategies using our complete Lab feature set:
\begin{itemize}
    \item \textbf{Simple MLP}: Flattens all features and maps them via a standard 3-layer MLP.
    \item \textbf{Bottleneck MLP}: Utilizes a bottleneck MLP architecture combined with base token mapping (identical to Features C in the previous section).
    \item \textbf{MR-Mapper (Ours)}: Employs our proposed architecture, which processes features through base, residual (delta), and global branches using 1D and 2D convolutions to preserve spatial topology.
\end{itemize}

As shown in Table \ref{tab:mapper_ablation_d}, the Simple MLP employs a naive flattening strategy, which destroys the structural information of the features and leads to the worst performance (a Dist. to Ideal of $0.286$). The Bottleneck MLP filters out redundant information to some extent by reducing dimensionality before mapping, yielding a performance improvement (Dist. to Ideal drops to $0.281$). In contrast, our MR-Mapper achieves the best generation quality (the Dist. to Ideal reaches an optimal $0.259$, and the Content Similarity significantly improves to $0.760$). More importantly, the Params across these three mapping strategies exhibits a progressively decreasing trend. Traditional MLPs introduce a large number of redundant parameters, whereas the MR-Mapper explicitly models residual differences and global statistics while preserving spatial topology via weight-sharing convolutions. Although the MACs are slightly higher than the Bottleneck MLP due to the sliding window mechanism of convolutions, MR-Mapper achieves the most efficient multi-dimensional feature aggregation with significantly fewer parameters (only $0.38$M), demonstrating its dual superiority in both performance and architectural efficiency.

\begin{table}[htbp]
    \centering
    \caption{Quantitative comparison of different condition mapping modules. The proposed MR-Mapper achieves better style transfer performance while requiring significantly fewer parameters.}
    \label{tab:mapper_ablation_d}
    \resizebox{\textwidth}{!}{
        \begin{tabular}{l|ccc|cc}
            \toprule
            Mapping Module   & Content Sim. $\uparrow$ & Style Sim. $\uparrow$ & Dist. to Ideal $\downarrow$ & Params (M) $\downarrow$ & MACs (M) $\downarrow$ \\
            \midrule
            Simple MLP       & 0.723                   & 0.866                 & 0.286                       & 428.97                  & 428.87                \\
            Bottleneck MLP   & 0.726                   & \textbf{0.871}        & 0.281                       & 5.24                    & \textbf{5.24}         \\
            MR-Mapper (Ours) & \textbf{0.760}          & 0.870                 & \textbf{0.259}              & \textbf{0.38}           & 13.82                 \\
            \bottomrule
        \end{tabular}
    }
\end{table}

\subsection{Selection of LUT Size}
The size of the 3D LUT, denoted as $D$, directly determines the precision of color mapping and the computational overhead of the model. To investigate the impact of the LUT size on the color transfer performance, we conduct an ablation study by setting $D \in \{8, 12, 16, 20, 24\}$.

As illustrated in Table \ref{tab:lut_size}, when the LUT size increases from $8$ to $16$, the model exhibits noticeable improvements in both content preservation and stylization capabilities. However, as $D$ continues to increase to $20$ or $24$, the changes in performance metrics become marginal, with no significant further gains. It is worth noting that the number of parameters and memory footprint of the 3D LUT grow cubically, with a spatial complexity of $\mathcal{O}(D^3 \times 3)$. Therefore, larger LUT sizes introduce drastically higher memory consumption and computational overhead without providing proportional performance benefits. Taking both visual quality and real-time inference efficiency into comprehensive consideration, we ultimately select $D=16$ as the hyperparameter setting for all experiments in this paper.

\begin{table}[htbp]
    \centering
    \caption{Quantitative comparison of different LUT sizes.}
    \label{tab:lut_size}
    \begin{tabular}{cccc}
        \toprule
        LUT Size ($D$) & Content Sim. $\uparrow$ & Style Sim. $\uparrow$ & Dist. to Ideal $\downarrow$ \\
        \midrule
        8              & 0.705                   & 0.834                 & 0.322                       \\
        12             & 0.706                   & 0.858                 & 0.303                       \\
        16             & 0.760                   & 0.870                 & 0.259                       \\
        20             & 0.752                   & 0.865                 & 0.268                       \\
        24             & 0.754                   & 0.874                 & 0.260                       \\
        \bottomrule
    \end{tabular}
\end{table}

\section{Broader Impacts}

\paragraph{Positive Impacts.} By leveraging the highly efficient 3D Look-Up Table (LUT) architecture, our stylization framework ensures low energy consumption and a reduced carbon footprint. This lightweight design aligns with Green AI goals and democratizes high-quality editing by enabling real-time processing on resource-constrained edge devices (e.g., mobile phones) without relying on heavy cloud infrastructure.

\paragraph{Negative Impacts and Mitigation.} While our method only adjusts low-level colors and luminance, it could inadvertently alter human skin tones, raising fairness and representation concerns regarding racial identity. Additionally, it could be misused to visually enhance fabricated content. To mitigate these risks, we recommend incorporating skin-tone preservation masks or fairness-aware constraints when processing human subjects. Furthermore, real-world deployments should adopt provenance tracking (e.g., C2PA standards) or digital watermarking to ensure media authenticity.

\end{document}